\documentclass[conference]{IEEEtran}
\IEEEoverridecommandlockouts
\usepackage{cite}
\usepackage{amsmath,amssymb,amsfonts}
\usepackage{algorithmic}
\usepackage{graphicx}
\usepackage{textcomp}
\usepackage{xcolor}

\usepackage{algorithm}
\usepackage{algorithmic}
\usepackage{multirow}
\usepackage{amsmath,amssymb,amsfonts}
\usepackage{booktabs}
\usepackage{multirow}
\usepackage{subfigure} 
\usepackage{newfloat}
\usepackage{listings}

\def\BibTeX{{\rm B\kern-.05em{\sc i\kern-.025em b}\kern-.08em
    T\kern-.1667em\lower.7ex\hbox{E}\kern-.125emX}}
\begin{document}

\title{Ask, Attend, Attack: A Effective Decision-Based Black-Box Targeted Attack for Image-to-Text Models\\
}

% \author{\IEEEauthorblockN{Anonymous Authors}}
\author{\IEEEauthorblockN{1\textsuperscript{st} Qingyuan Zeng}
\IEEEauthorblockA{\textit{Institute of Artificial Intelligence} \\
\textit{Xiamen University}\\
Fujian, China \\
36920221153145@stu.xmu.edu.cn}
\and
\IEEEauthorblockN{2\textsuperscript{nd} Zhenzhong Wang}
\IEEEauthorblockA{\textit{Department of Computing} \\
\textit{The Hong Kong Polytechnic University}\\
Hongkong, China \\
zhenzhong16.wang@connect.polyu.hk}
\and
\IEEEauthorblockN{3\textsuperscript{rd} Yiu-ming Cheung}
\IEEEauthorblockA{\textit{Department of Computer Science} \\
\textit{Hong Kong Baptist University}\\
Hongkong, China \\
ymc@comp.hkbu.edu.hk}
\and
\IEEEauthorblockN{4\textsuperscript{th} Min Jiang*}
\IEEEauthorblockA{\textit{School of Informatics} \\
\textit{Xiamen University}\\
Fujian, China \\
minjiang@xmu.edu.cn}
\thanks{The corresponding author: Min Jiang, minjiang@xmu.edu.cn}
}

\maketitle

\begin{abstract}
While image-to-text models have demonstrated significant advancements in various vision-language tasks, they remain susceptible to adversarial attacks. Existing white-box attacks on image-to-text models require access to the architecture, gradients, and parameters of the target model, resulting in low practicality. Although the recently proposed gray-box attacks have improved practicality, they suffer from semantic loss during the training process, which limits their targeted attack performance. To advance adversarial attacks of image-to-text models, this paper focuses on a challenging scenario: decision-based black-box targeted attacks where the attackers only have access to the final output text and aim to perform targeted attacks. Specifically, we formulate the decision-based black-box targeted attack as a large-scale optimization problem. To efficiently solve the optimization problem, a three-stage process \textit{Ask, Attend, Attack}, called \textit{AAA}, is proposed to coordinate with the solver. \textit{Ask} guides attackers to create target texts that satisfy the specific semantics. \textit{Attend} identifies the crucial regions of the image for attacking, thus reducing the search space for the subsequent \textit{Attack}. \textit{Attack} uses an evolutionary algorithm to attack the crucial regions, where the attacks are semantically related to the target texts of \textit{Ask}, thus achieving targeted attacks without semantic loss. Experimental results on transformer-based and CNN+RNN-based image-to-text models confirmed the effectiveness of our proposed \textit{AAA}.
\end{abstract}

\section{Introduction}

Image-to-text models, referring to generating descriptive and accurate textual descriptions of images, have received increasing attention in various applications, including image-captioning~\cite{li2022blip,li2023blip2}, visual-question-answering~\cite{antol2015vqa,kim2021vilt}, and image-retrieval~\cite{cao2018vggface2,lu2019vilbert}. Despite the remarkable progress, they are vulnerable to deliberate attacks, giving rise to concerns about the reliability and trustworthiness of these models in real-world scenarios. For example, one may mislead models to output harmful content such as political slogans and hate speech by making imperceptible perturbations to images~\cite{chen2018attacking,lapid2023iseedeadpeople,zhao2023onevaluating}. 
\begin{figure}[t]
\centering
\includegraphics[width=1\columnwidth]{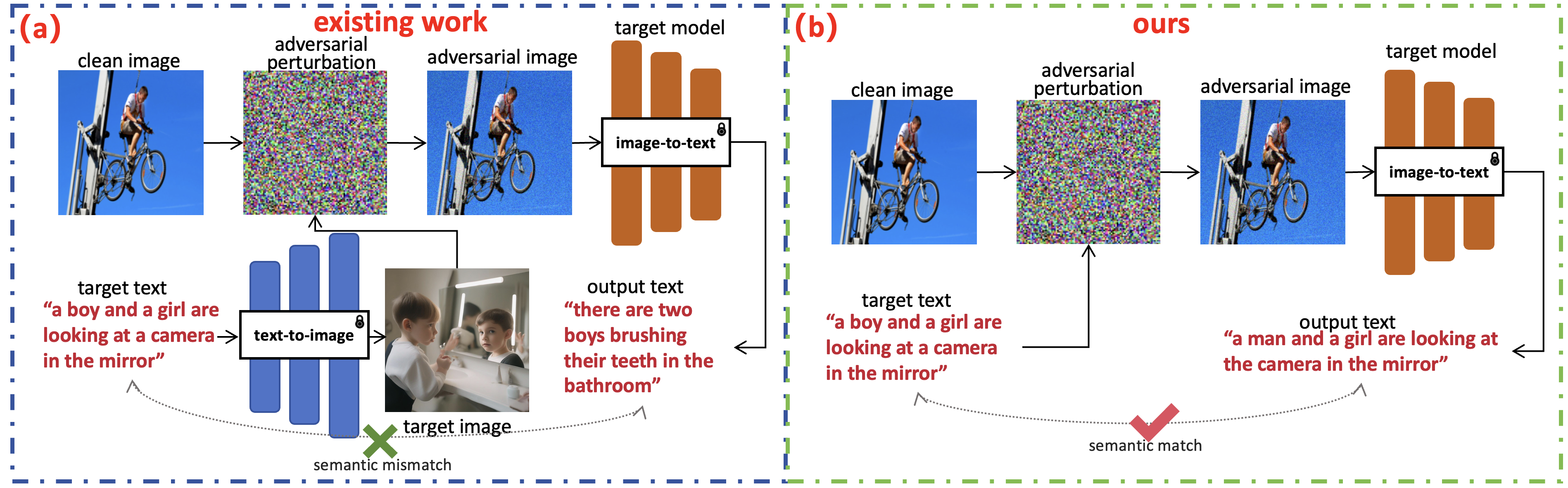}
\caption{The semantic loss problem existing in existing gray-box targeted attack methods.}
\label{fig:whyWeak }
\end{figure}

To gain insight into the reliability and trustworthiness of the image-to-text models, a series of adversarial attack methods have been proposed to poison the outputted textual descriptions of given images~\cite{chen2018attacking,kwon2022restricted,lapid2023iseedeadpeople,zhao2023onevaluating}. Specifically, based on the attacker’s level of access to information about the target model, they can be divided into three categories: white-box attacks~\cite{chen2018attacking,bhattad2020unrestricted,kwon2022restricted}, gray-box attacks~\cite{lapid2023iseedeadpeople,zhao2023onevaluating}, and black-box attacks~\cite{dong2019efficient,shi2022query}. The white-box attacks can obtain target models' information including the entire architecture, parameters, gradients of both the image encoder and text decoder, and probability of each word of the output text. Gray-box attacks can only access the architecture, parameters, and gradients of the image encoder, while black-box attacks cannot access any internal information of the target model, but only the output text of the model. Furthermore, black-box attacks can be divided into score-based and decision-based attacks. Score-based black-box attacks can access the probability of each word of the output text~\cite{shi2022query}, while decision-based black-box attacks can only access the output text~\cite{dong2019efficient,jia2021iou,jiang2023efficient}. Because less information about the target models is provided, decision-based black-box attacks are more challenging than other categories~\cite{jiang2023efficient}. Additionally, these attack methods can be categorized based on whether the attacker is able to specify the incorrect output text, dividing them into two types: targeted and untargeted attacks ~\cite{wu2022learning,lapid2023iseedeadpeople}.

% On the other hand, these approaches can also be classified into two classes: targeted attack and untargeted attack, i.e., whether the attacker can specify the wrong output text~\cite{wu2022learning,lapid2023iseedeadpeople}.

Although numerous adversarial attack methods for image-to-text models have been proposed, to our best knowledge, the study on black-box attacks is under-explored, especially decision-based black-box targeted attacks. This kind of attack is more challenging due to the following reasons. Firstly, less information on the target model can be accessed. Specifically, only the output text instead of gradients, architectures, parameters, and the probability of each word in the output text is available. Secondly, the attackers not only cause the target model to output incorrect text, but also outputs the specified target text. Existing attacks easily suffer from the loss of semantics, resulting in the inability to effectively output the specified target text. Figure \ref{fig:whyWeak } (a) show that transfer+query ~\cite{zhao2023onevaluating} fabricates one target text to poison the target image-to-text model, leading to this model outputting an incorrect text. However, the output text could mismatch the original semantics of the target text, as the target image-to-text model may focus on secondary information while ignoring the crucial semantics of the target text behind the target image, resulting in semantic loss. More examples are in Appendix \ref{sec:semantic loss example}.

% See Appendix \ref{sec:semantic loss example} for more analysis and examples on semantic loss.

To narrow the research gap, we propose a decision-based black-box targeted attack approach for image-to-text models. In our work, only the output text of the target model can be accessed, which is closer to the real-world cases~\cite{dong2019efficient}. Additionally, Figure \ref{fig:whyWeak } (b) demonstrates our targeted attack method, which optimizes against the target text directly under the decision-based black-box conditions, preventing semantic loss and maintaining semantic consistency with the target text.

\begin{figure}[t]
\centering
\includegraphics[width=0.8\columnwidth]{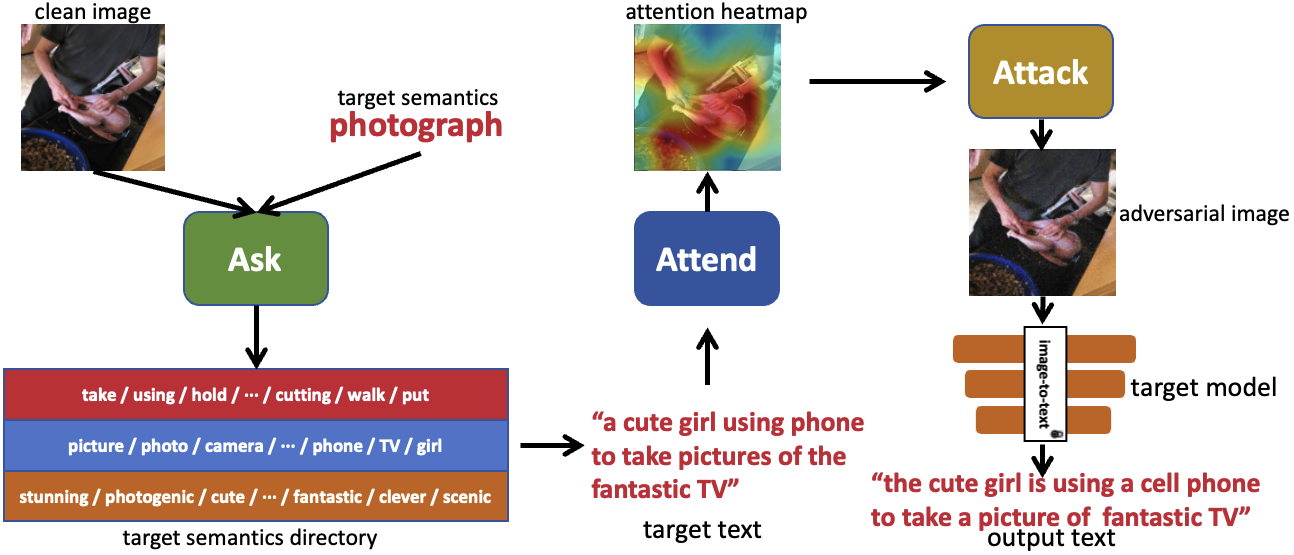}
\caption{Diagram of our decision-based black-box targeted attack method \textit{Ask, Attend, Attack}.}
\label{fig:framework}
\end{figure}

Perturbing pixels in the image can change the output text. Therefore, the objective of the targeted attack can be considered to find the imperceptible pixel modification to make the output text similar to the target text. In this manner, the targeted attack can be formulated as a large-scale optimization problem, where pixels are decision variables and the optimization objective is to poison the output text. Inspired by the distinctive competency of evolutionary algorithms for solving large-scale optimization problems ~\cite{omidvar2013cooperative,9552479,hong2023improving}, we develop a dedicated evolutionary algorithm-based framework for decision-based black-box targeted attacks on image-to-text models. However, directly applying evolutionary algorithms to solve this large-scale optimization problem could suffer from low search efficiency, due to the numerous pixels and their wide range of values.  To address the issue, we embed three-step processes, i.e., \textit{Ask, Attend, Attack}, into the proposed evolutionary algorithm-based attacks. As shown in Figure \ref{fig:framework}, during the \textit{Ask} stage, attackers can arbitrarily specify words related to certain semantics, such as \textit{photograph}. Then, candidate words (e.g., \textit{camera}, \textit{scenic}, and \textit{phone}) that are related to certain semantics are searched. Meanwhile, these words are close to the clean image in the feature space of the target image-to-text model. By selecting words from the candidate words, the target text (e.g., \textit{a cute girl using a phone to take pictures of the fantastic TV}) related to the attacker's specified semantics can be formed to poison the target model. Subsequently, based on the attention mechanism, \textit{Attend} identifies the crucial regions of the clean image (e.g. attention heatmap), thus reducing the search space for the subsequent \textit{Attack}. Lastly, \textit{Attack} uses a differential evolution strategy to impose imperceptible adversarial perturbations to the crucial regions, where the optimization objective is to minimize the discrepancy between the target text in \textit{Ask} stage and the output text of the target model. Our contributions can be summarized as follows:

\begin{enumerate}
    \item We first propose a decision-based black-box targeted attack \textit{Ask, Attend, Attack} (\textit{AAA}) for image-to-text models. Specifically, our method achieves targeted attacks without losing semantics while only the model's output text can be accessed.
    \item We designed a target semantic directory to guide attackers in creating target text and utilized attention heatmaps to significantly reduce search space. This improves the search efficiency of evolutionary algorithms in adversarial attacks and makes attacks difficult to perceive.
    % \item We use the target semantic directory and target text attention heatmap to accelerate the search efficiency of the designed evolutionary algorithm\zhenzhong{poor concealment in adversarial attacks?}. 
    \item We conducted extensive experiments on the Transformer-based VIT-GPT2 model and CNN+RNN-based Show-Attend-Tell model, which are the two most-used image-to-text models in HuggingFace, and surprisingly found that our decision-based black-box method has stronger attack performance than existing gray-box methods.
\end{enumerate}

\section{Related work}

\subsection{White-box Attack}
In white-box attacks, the attacker has full access to all parameters, gradients, architecture of the target model, and the probability of each word of the output text. The authors in ~\cite{chen2018attacking} add invisible perturbations to the image to make the image-to-text model produce wrong or targeted text outputs.  The authors in ~\cite{xu2018fooling} add global or local perturbations to the image to make the vision and language models unable to correctly locate and describe the content of the image. The authors in ~\cite{bhattad2020unrestricted} modify the content of the image at the semantic level to make the image-to-text model output text that is inconsistent with the original image. The authors in ~\cite{zhang2020fooled} crafts adversarial examples with semantic embedding of targeted captions as perturbation in the complex domain. The authors in ~\cite{ji2020attacking} preserves the accuracy of non-target words while effectively removing target words from the generated captions. The authors in ~\cite{huang2021igseg} generate coherent and contextually rich story endings by integrating textual narratives with relevant visual cues. The authors in ~\cite{kwon2022restricted} add limited-area perturbations to the image to make the image-to-text model fail to correctly describe the content of the perturbed area. The above methods require complete information of the image-to-text target model, including architecture, gradients, parameters, and probability distribution of the output text, which limits their practicality.

\subsection{Gray-box Attack}
To improve the practicality of adversarial attacks for image-to-text models, recent research explores how to attack with partial knowledge of the target model. All existing gray-box attack studies ~\cite{chaturvedi2020mimicandfool,wu2022learning,lapid2023iseedeadpeople,zhao2023onevaluating} assume full access to the image encoder of the image-to-text model.  The basic idea of gray-box targeted attacks is to reduce the distance between the adversarial image and the target image generated based on the target text in the image encoder’s feature space. 
The authors in ~\cite{chaturvedi2020mimicandfool} generates adversarial images to mimic the feature representation of original images. The authors in ~\cite{wu2022learning} use a generative model to destroy the image encoder’s features, achieving the untargeted attack. The authors in ~\cite{lapid2023iseedeadpeople} minimize the feature distance in the image encoder between the adversarial image and the target image, thereby using gradient back-propagation to optimize the adversarial image and achieve the targeted attack. The authors in ~\cite{zhao2023onevaluating} combine existing gray-box method ~\cite{lapid2023iseedeadpeople} with pseudo gradient estimation  method ~\cite{nesterov2017random} to achieve better performance in targeted attack. It is worth noting that they ~\cite{zhao2023onevaluating} call their method a black-box attack, but since they use the image encoder of the target model as the surrogate model, we classify their method as a gray-box attack. These gray-box attacks on image-to-text models are more practical than white-box attacks, but it is still unrealistic to assume that attackers can access the image encoder of the image-to-text model. Moreover, existing gray-box methods may have poor targeted attack performance due to the semantic loss mentioned above.

\section{Methodology}

\subsection{Problem Formulation}
The image-to-text model $\mathcal{G}:\mathcal{X}\rightarrow\mathcal{Y}$ maps the image domain $\mathcal{X}$ to the text domain $\mathcal{Y}$. A well-trained model should be able to accurately describe the content of the image using grammatically correct and contextually coherent text. Given a target text $y_t$, the attacker's goal is to find an adversarial image $\mathbf{x_{adv}}$ that is visually similar to clean image $\mathbf{x}$ and can generate an adversarial text $y_{adv}$ that is semantically similar to $y_t$. We formalize the optimization problem for black-box targeted attack as:
\begin{equation}
\begin{aligned}
\arg\max_{\mathbf{x}_{adv}}   S(\mathcal{G}(\mathbf{x}_{\mathbf{adv}}),y_t) \ \ 
\text{s.t.} \;  \frac{1}{n}\sum_{i=1}^{n}\|\mathbf{x_{adv}}(i) - \mathbf{x}(i)\|  \leq \epsilon,
\end{aligned}
\end{equation}
where $S(\cdot,\cdot)$ represents the semantic similarity function between two texts,  $\epsilon$ is the threshold for the average perturbation size per pixel,  $\mathbf{x_{adv}}(i)$ and $\mathbf{x}(i)$ represents the value of the $i$-th pixel in the adversarial and clean images. $n$ is the total number of pixels in all channels of the image.

\subsection{Overview}
To enhance the efficiency and stealth of decision-based black-box attacks, we propose the \textit{Ask, Attend, Attack} (\textit{AAA}) framework as shown in Figure \ref{fig:framework}. \textit{Ask}: We compile a semantic dictionary from words within the input image’s search space that align with the attacker’s specified semantics. This facilitates targeted text generation, meeting the attacker’s target semantics while simplifying the search process. \textit{Attend}: We employ attention visualization and a surrogate model to generate an attention heatmap for the target text on the image, narrowing the search to significant decision variables and enhancing perturbation stealth. \textit{Attack}: We use the differential evolution in the reduced search space to find the optimal solution that can mislead the target model to output target text. The framework’s pseudo-code is detailed in Appendix ~\ref{sec:pseudo_code}.

% In order to conduct a more efficient and covert  decision-based black-box targeted attack, we propose the \textit{Ask, Attend, Attack} (\textit{AAA}) as shown in Figure \ref{fig:framework}. \textit{Ask}: We find words (nouns, adjectives, verbs) that match the target semantics specified by the attacker in the spherical search space of the input image to form a target semantic dictionary. Using words in the dictionary to create target text can not only satisfy the target semantics specified by the attacker, but also greatly reduce the difficulty of searching. \textit{Attend}: we use attention visualization technology and a surrogate model to calculate the attention heatmap of the target text on the image. By reducing the search range of unimportant decision variables (pixels), we can improve the search efficiency and the concealment of adversarial perturbation. \textit{Attack}: We use the differential evolution in the reduced search space to find the optimal solution that can mislead the target model to output target text. We summarize our framework in pseudo-code in Appendix ~\ref{sec:pseudo_code}.

\subsection{Ask Stage}

According to the target semantics, the goal of \textit{Ask} is to find words in the feature space of the target model to form a target semantic dictionary. These words should be closer to the input image. Firstly, we treat each pixel in each channel of image $\mathbf{x}$ as a variable, which means the search space size is the product of length, width, and number of channels. And then generate NP (number of population) individuals to form a population based on the following formula:
\begin{equation}
\begin{aligned}
\mathbf{x}_j(i) = \mathbf{x}(i) + rand(-1, 1) \cdot \eta,
\end{aligned}
\end{equation}
where $\mathbf{x}(i)$ is the $i$-th variable of clean image $\mathbf{x}$, $\mathbf{x}_j(i)$  is the $i$-th variable of the $j$-th individual in the population, $\eta$ is a hyperparameter about the maximum search range, \textit{rand}(-1,1) is a random number from the range of -1 to 1.

Secondly, for each variable, random mutation occurs between different individuals. The mutation for the $i$-th variable of the $j$-th individual $\mathbf{x}_j(i)$ is as follows:
\begin{equation}
\begin{aligned}
\mathbf{v}^g_j(i) = \mathbf{x}^g_{r1}(i) + F*(\mathbf{x}^g_{r2}(i) - \mathbf{x}^g_{r3}(i) ),
\end{aligned}
\end{equation}
where $\mathbf{v}^g_j(i)$ is the mutated variable for mutation in the $g$-th generation of $\mathbf{x}_j(i)$.  $\mathbf{x}^g_{r1}(i)$, $\mathbf{x}^g_{r2}(i)$, and $\mathbf{x}^g_{r3}(i)$ are three randomly selected individuals from the current population who are different from each other, $F$ is the scaling factor.

Thirdly, each individual crossovers with the mutated individuals with a certain probability of generating candidate individuals. The formula is as follows:
\begin{equation}
\begin{aligned}
\mathbf{u}^g_{j}(i)=\left\{\begin{array}{lc}
\mathbf{v}^g_{j}(i), & \text { if rand }(0,1) \leq CR,  \\
\mathbf{x}^g_{j}(i), & \text { otherwise, }
\end{array}\right.
\end{aligned}
\label{crossover}
\end{equation}
where $CR$ is crossover probability factor, $\mathbf{u}^g_{j}(i)$ is the $i$-th variable of the candidate individual in the $g$-th generation of the $j$-th individual in the population.

Fourthly, we use WordNet ~\cite{banerjee2005meteor}, a synonym dictionary, to measure the similarity between the target semantics and each individual’s output text. WordNet groups words with the same semantics into synonyms, each representing a basic concept. We use WordNet to count the same semantic words $m$ between each individual's output text and the target semantics. We calculate the $Precison$=($m/t$) and $Recall$=($m/r$), where $t$ is the output text word count and $r$ is the target semantics word count. Then, we calculate semantic similarity using the following formula:
\begin{equation}
\begin{aligned}
S_{sem} =  \frac{(1 - \gamma (\frac{ch}{m})^\theta)(\alpha^2 + 1) \cdot Precision \cdot Recall }{\alpha^2\cdot Precision + Recall},
\end{aligned}
\end{equation}
where $S_{seg}$ is the semantic similarity between the individual's output text and the target semantics ~\cite{banerjee2005meteor}, $\alpha$ balances the precision and recall weights, $\gamma$ and $\theta$ control the penalty factor strength, $ch$ is the number of consecutive word sets that match between the output text and the target semantics, with fewer chunks meaning more consistent word order.

Ultimately, we select offspring based on $S_{sem}$, choosing the current and candidate individuals that match the target semantics better as the next generation:
\begin{equation}
\begin{aligned}
\mathbf{x}^{g+1}_{j}=\left\{\begin{array}{lc}
\mathbf{u}^g_{j}, & S_{sem}(\mathcal{G}(\mathbf{u}^g_{j}), TS) \geq S_{sem}(\mathcal{G}(\mathbf{x}^g_{j}), TS), \\
\mathbf{x}^g_{j}, & \text { otherwise,}
\end{array}\right.
\end{aligned}
\end{equation}
where $TS$ is the attacker's target semantics. We extract nouns, adjectives, and verbs from the output texts of all the more semantically relevant and preserved individuals of each generation, expanding the target semantic dictionary. We use $\mathcal{G}(\mathbf{x}^{g+1}_{j}) = \{w_1,w_2,\cdot\cdot\cdot,w_n\}$  to represent the target model $\mathcal{G}$'s output text for the next generation of individuals, where $w_i$ is the $i$-th word of the text and $n$ is the word count. Then we use the following formula to extract important words and make  a dictionary:
\begin{equation}
\begin{aligned}
\mathbf{D}^{g+1}_j = \{ w \in \mathcal{G}(\mathbf{x}^{g+1}_{j}) \ | \ \text{$w$ is noun, adjective or verb}\},
\end{aligned}
\end{equation}
where  $\mathbf{D}^{g+1}_j$ is the dictionary for the preserved individual. We combine the dictionaries of each preserved individual in each generation to get the target semantic dictionary $\mathbf{D} =  \mathbf{D}^{2}_1 \cup \mathbf{D}^{2}_2 \cup \cdot\cdot\cdot \cup \mathbf{D}^{m}_{\text{NP}} $, where $m$ is the total number of generation. The attacker selects words from dictionary $\mathbf{D}$ that match the specified semantics to make the target text $y_t$. Words in dictionary $\mathbf{D}$ near input image $\mathbf{x}$ in feature space enhance searchability, enabling more efficient targeted attacks.

% Because the words in dictionary $\mathbf{D}$ are close to the input image $\mathbf{x}$ in the feature space and easy to search, attackers can do the targeted attack with higher efficiency.

\subsection{Attend Stage}
The goal of \textit{Attend} is to calculate the target text's attention area on the image  $\mathbf{x}$.  Because we do not have access to the internal information of the target model, we can only calculate the Grad-CAM attention heatmap ~\cite{selvaraju2017grad} with the help of surrogate model $f$ (such as ResNet trained in ImageNet). The surrogate model’s sole purpose is to the compute attention heatmap. Since different models produce similar heatmaps for the same target text and input image, selecting a well-established visual model suffices  ~\cite{wang2021dual}. The calculation formula of attention heatmap $\mathbf{A}$ is as follows:
\begin{equation}
    \mathbf{A}(i, j)=\operatorname{MAX}\left(0,  \frac{1}{Z}\sum_{k}   \sum_{i} \sum_{j}  \cdot \frac{\partial y^{c^*}}{\partial \mathcal{F}_{k}(i,j)} \cdot \mathcal{F}_{k}(i,j)\right),  
\label{equation:gradcam}
\end{equation}
where $\mathbf{A}(i, j)$ is the decision-making contribution of the image to the target text at pixel $(i,j)$, $\mathcal{F}_{k}(i,j)$ is the pixel  $(i,j)$ of the feature map of the $k$-th convolution kernel of the last convolutional layer of the surrogate model $f$,   $Z$ is the feature map’s pixel count, $y^{c^*}$ is the probability that $f$ predicts that the image $\mathbf{x}$ belongs to class $c^*$. We use $\mathbf{C}=\{ c_1, c_2, \cdot\cdot\cdot, c_{1000} \}$ for the ImageNet category names, where $c_i$ is the $i$-th category name. We make the category text $y_{c_i} = ``\textit{a \ photo \ of}" + c_i$ from the category name $c_i$. We calculate the category $c^*$ as:
\begin{equation}
\begin{aligned}
c^* = \underset{c_i \in \mathbf{C}}{\operatorname{argmax}} \ \frac{E(y_t) \cdot  E(y_{c_i})} { \| E(y_t) \|_2 \| E(y_{c_i}) \|_2 }, 
\end{aligned}
\label{classC}
\end{equation}
where $E$ is the text encoder of the pre-trained CLIP model, and $c^*$ is the closest category to the target text. We substitute $c^*$ into Formula \ref{equation:gradcam} to get the target text’s attention heatmap $\mathbf{A}$. $\mathbf{A}(i, j)$ is the pixel $(i, j)$'s contribution to the target text. 1 means more contribution, and 0 means less contribution.

\begin{table*}[htbp]\tiny
  \centering
  \caption{Performance comparison (\%) of different attack methods.}
    \begin{tabular}{c|c|cccc|cccc}
    \toprule
 \multicolumn{1}{c|} {\multirow{2}[2]{*}{$\epsilon$}} & \multirow{2}[2]{*}{Attack Methods} & \multicolumn{4}{c|}{VIT-GPT2} & \multicolumn{4}{c}{Show-Attend-Tell}  \\
    
\cmidrule{3-10}

& & METEOR & BLEU & CLIP & SPICE  & METEOR & BLEU & CLIP & SPICE   \\

    \midrule
                         & Clean Sample & 0.201±0.11 & 0.24±0.11  & 0.64±0.07 & 0.156±0.07 & 0.21±0.11 & 0.229±0.13 & 0.646±0.09 & 0.179±0.08 \\
    \midrule
                        & transfer (black) & 0.206±0.11 & 0.246±0.11  & 0.639±0.07 & 0.165±0.07 & 0.211±0.12 & 0.225±0.14 & 0.648±0.09 & 0.185±0.11 \\
                        & transfer+query (black) & 0.221±0.16 & 0.264±0.15  & 0.651±0.18 & 0.167±0.07 & 0.219±0.11 & 0.231±0.14 & 0.654±0.05 & 0.187±0.14 \\
\multirow{2}[1]{*}{25} & transfer (gray) & 0.414±0.23 & 0.396±0.14  & 0.821±0.09 & 0.32±0.16 & 0.382±0.26 & 0.348±0.17 & 0.782±0.11 & 0.299±0.17  \\
                        & transfer+query (gray) & 0.433±0.21 & 0.411±0.12  & 0.832±0.13 & 0.35±0.09 & 0.401±0.21 & 0.355±0.15 & 0.794±0.11 & 0.311±0.13 \\
                        & \textit{AAA} (w/o \textit{Attend}) & 0.541±0.25 & 0.519±0.19 & 0.854±0.24 & 0.477±0.11 & 0.642±0.19 & 0.564±0.19 & 0.841±0.06 & 0.455±0.14 \\
                        & \textit{AAA} (w/o \textit{Ask}) & 0.398±0.21 & 0.384±0.18  & 0.795±0.25 & 0.412±0.13 & 0.364±0.21 & 0.322±0.19 & 0.754±0.08 & 0.376±0.13 \\
                        & \textit{AAA} & \textbf{0.696±0.21} & \textbf{0.658±0.22}  & \textbf{0.952±0.29} & \textbf{0.634±0.15} & \textbf{0.855±0.15} & \textbf{0.799±0.21}  & \textbf{0.964±0.04} & \textbf{0.786±0.14} \\
    \hline
                        & transfer (black) & 0.204±0.09 & 0.241±0.15  & 0.627±0.18 & 0.164±0.07 & 0.232±0.13 & 0.236±0.14 & 0.643±0.08 & 0.187±0.09 \\
                        & transfer+query (black) & 0.211±0.14 & 0.256±0.15  & 0.644±0.15 & 0.181±0.09 & 0.245±0.13 & 0.246±0.11 & 0.656±0.06 & 0.203±0.09 \\
\multirow{2}[1]{*}{15} & transfer (gray) & 0.398±0.24 & 0.381±0.15 & 0.816±0.11 & 0.325±0.16 & 0.361±0.24 & 0.359±0.17 & 0.778±0.11 & 0.296±0.16 \\
                        & transfer+query (gray) & 0.408±0.19 & 0.399±0.11  & 0.824±0.15 & 0.341±0.13 & 0.375±0.19 & 0.368±0.15 & 0.784±0.11 & 0.311±0.13 \\
                        & \textit{AAA} (w/o \textit{Attend}) & 0.461±0.21 & 0.423±0.15 & 0.808±0.11 & 0.375±0.09 & 0.438±0.15 & 0.434±0.16 & 0.827±0.04 & 0.422±0.14 \\
                        & \textit{AAA} (w/o \textit{Ask}) & 0.378±0.25 & 0.361±0.17  & 0.768±0.15 & 0.356±0.15 & 0.341±0.15 & 0.337±0.18 & 0.749±0.07 & 0.365±0.13 \\
                        & \textit{AAA} & \textbf{0.556±0.31} & \textbf{0.504±0.26}  & \textbf{0.851±0.12} & \textbf{0.44±0.17} & \textbf{0.617±0.25} & \textbf{0.574±0.22} & \textbf{0.913±0.05} & \textbf{0.553±0.14} \\

    \bottomrule
    \end{tabular}%
  \label{tab:total}%
\end{table*}%

\subsection{Attack Stage}

The goal of \textit{Attack} is to search for the best individual (adversarial sample) that  outputs the target text $y_t$ in the smaller search space reduced by the attention heatmap. Firstly, we copy the attention heatmap $\mathbf{A}$ three times in the channel dimension to match the shape of the image $\mathbf{x}$. We generated NP (number of population) individuals as a population with this formula:
\begin{equation}
\begin{aligned}
\mathbf{x}_j(i) = \mathbf{x}(i) + rand(-\mathbf{A}(i), \mathbf{A}(i)) \cdot \eta, 
\end{aligned}
\end{equation}
where $\mathbf{x}(i)$ is the $i$-th variable of clean image $\mathbf{x}$,  $\mathbf{x}_j(i)$ is the $i$-th variable of the $j$-th individual in the population, $\mathbf{A}(i)$  is the contribution of the $i$-th variable to the target text, and $rand(-\mathbf{A}(i), \mathbf{A}(i))$ is a random number in the range from $-\mathbf{A}(i)$ to $\mathbf{A}(i)$. The value of $\mathbf{A}$ is less than 1, and its mean and median are about [0.3,0.4]. The search space volume from the attention heatmap is much smaller than a hypersphere with radius $\eta$, because the radius and volume have an exponential relationship. This improves the search efficiency and concealment of adversarial perturbation.

Secondly, in order to accelerate convergence and better find the global optimal solution, we use the following CurrentToBest mutation ~\cite{zhang2009jade}:
\begin{equation}
\begin{aligned}
\mathbf{v}^g_j(i) = \mathbf{x}^g_{j}(i) & + F*( \mathbf{x}^g_{r1}(i) - \mathbf{x}^g_{r2}(i)) \\ 
 & + F*( \mathbf{x}^g_{best}(i) - \mathbf{x}^g_{j}(i)), \\
\end{aligned}
\end{equation}
where $\mathbf{x}^g_{j}(i)$ is the $i$-th variable of the $j$-th individual in the $g$-th generation, $\mathbf{v}^g_j(i)$ is the mutated variable, $\mathbf{x}^g_{best}$ is the best fitness individual in the $g$-th generation population, $\mathbf{x}^g_{r1}$ and $\mathbf{x}^g_{r2}$ are two randomly selected individuals in the $g$-th generation population, and $F$ is the scaling factor. The main advantage of this mutation strategy is that it combines the information of the current individual $\mathbf{x}_{j}$ and the best fitness individual $\mathbf{x}_{best}$, which can better guide the search process towards the direction of the optimal solution.

Thirdly, we use Formula \ref{crossover} to calculate the candidate individual $\mathbf{u}^g_{j}(i)$.  We design the following formula to calculate the deep feature similarity $S_{clip}$ between two texts ($u$ and $v$):
\begin{equation}
\begin{aligned}
S_{clip} = 1 -  \frac{E(u) \cdot  E(v)} { \| E(u) \|_2 \| E(v) \|_2 }, 
\end{aligned}
\label{eqa:CLIP}
\end{equation}
where $E$ is the text encoder of the pre-trained CLIP model. Text is discrete and complex, so it cannot calculate the distance directly ~\cite{jiang2019semantic}. Therefore, we use the CLIP text encoder $E$ to extract the deep features of the texts, and then calculate the feature distance to obtain the similarity $S_{clip}$ between the texts. The closer $S_{clip}$ is to 0, the higher the similarity between the two texts $u$ and $v$. 

Ultimately, we select offspring using the following formula:
\begin{equation}
\begin{aligned}
\mathbf{x}^{g+1}_{j}=\left\{\begin{array}{lc}
\mathbf{u}^g_{j}, & S_{clip}(\mathcal{G}(\mathbf{u}^g_{j}), y_t) \leq S_{clip}(\mathcal{G}(\mathbf{x}^g_{j}), y_t), \\
\mathbf{x}^g_{j}, & \text { otherwise, }
\end{array}\right.
\end{aligned}
\end{equation}
where $\mathbf{x}^{g+1}_{j}$ is the next individual with closer feature distance between the output text and the target text $y_t$. After performing the above evolutionary calculations multiple times, the optimal solution (adversarial sample) for outputting the target text is found.

\begin{figure}[t]
\centering
\includegraphics[width=1\columnwidth]{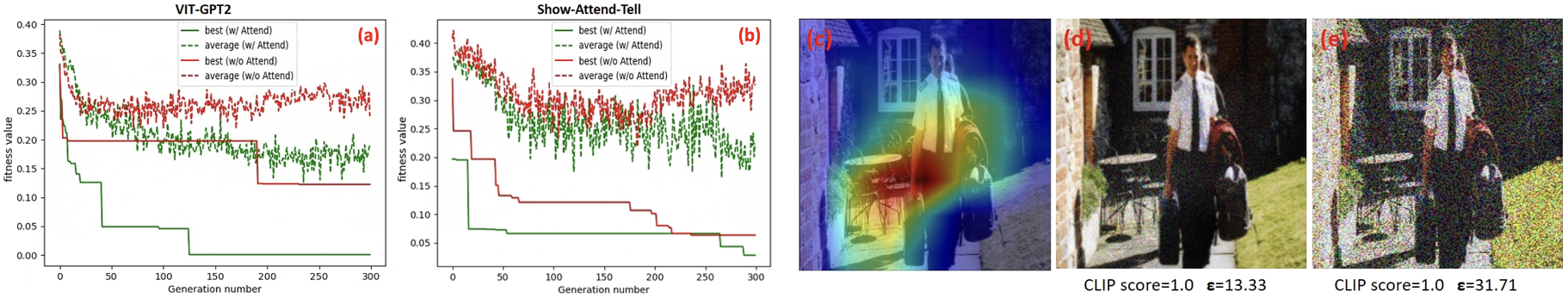}
\caption{We compared the convergence curves of populations with and without \textit{Attend} under the same perturbation size $\epsilon$ in (a-b). The fitness function is $S_{clip}$ in Formula \ref{eqa:CLIP}, where lower values mean stronger attacks. The dashed line is the average fitness value, and the solid line is the best fitness value. The green line is \textit{AAA} and the red line is \textit{AAA w/o Attend}. (c) shows the attention heatmap. (d) and (e) show the visual effects of adversarial image with and without $\textit{Attend}$, with minimal perturbation of 100\% attack success rate.}
\label{fig:attention}
\end{figure}

\section{Evaluation and Results}

\subsection{Experiment setups}
\paragraph{Model and dataset}

We experimented with the two most-used image-to-text models on HuggingFace: VIT-GPT2 (Transformer-based) ~\cite{nlpconnect2022} and Show-Attend-Tell (CNN+RNN-based)  ~\cite{Show2015}. VIT-GPT2 was trained on ImageNet-21k. Show-Attend-Tell was trained on MSCOCO-2014. We only used the target model’s output text, not its internal information like gradients, parameters, or word probability. Following this work ~\cite{lapid2023iseedeadpeople}, we used Flick30k as our dataset, which has 31783 images and 5 caption texts each. We removed samples with less than 0.7 similarities  between predicted text and truth text to ensure the target model’s accuracy on clean images.

\paragraph{Evaluation metrics}

We used these evaluation metrics in our experiments: (1) BLEU(\#4), an early machine translation metric that measures text precision ~\cite{papineni2002bleu}. 1 means similar, and 0 means dissimilar. (2) METEOR, a more comprehensive metric that considers synonyms, stems, word order, etc ~\cite{banerjee2005meteor}. 1 means similar, and 0 means dissimilar. (3) CLIP, the distance between the CLIP text encoder’s deep features for two texts ~\cite{radford2021learning}. 1 means similar, and 0 means dissimilar. (4) SPICE, an evaluation metric tailored for image-to-text models ~\cite{anderson2016spice}. 1 means similar, and 0 means dissimilar. (5) $\epsilon$, the mean perturbation size of each pixel of the adversarial sample ~\cite{lapid2023iseedeadpeople}. 

% Smaller value means higher concealment of adversarial perturbation.

% (4) iteration, the number of iterations for the differential evolution algorithm in \textit{Attack} to find the optimal solution (no more fitness convergence). Fewer iterations mean fewer queries and faster attack. (5) $\epsilon$, the mean perturbation size of each pixel of the adversarial sample. Smaller value means higher concealment of adversarial perturbation. (6) diversity, the number of words in the target semantic dictionary from \textit{Ask}. More words mean more diversity. (7) correlation, the average CLIP score between each word in the target semantic dictionary and the target semantics. The higher correlation, the more relevant the words in the target semantic dictionary are to the target semantics.

% \paragraph{Basic setups}
% We set the population size NP to 40, scaling factor $F$ to 0.7, cross probability factor $CR$ to 0.7, $\gamma$ to 0.5, $\alpha$ to 1, and $\theta$ to 3, and $\eta$ to $\epsilon$ required in the experiment divided by the average of attention heatmap $\mathbf{A}$.  Our device uses three GPUs of RTX2080ti with 11GB memory.

% and a CPU of Intel(R) Core(TM) i5-10400F.

\begin{figure}[t]
\centering
\includegraphics[width=1\columnwidth]{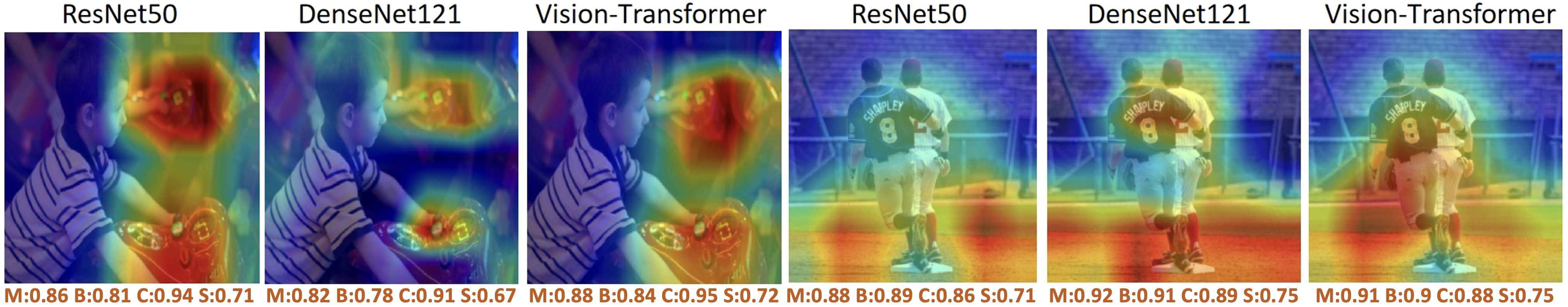}
\caption{Grad-CAM attention heatmaps of different surrogate models for the same target text \textit{a woman is holding a pair of shoes}. M is METEOR, B is BLEU, C is CLIP, S is SPICE.}
\label{fig:gradcam_backbone}
\end{figure}

\subsection{Experiment results}

\paragraph{Comparison experiment of existing gray-box attacks.}

We evaluate state-of-the-art gray-box attacks ~\cite{lapid2023iseedeadpeople,zhao2023onevaluating} on image-to-text models. We designate the gray-box attack ~\cite{lapid2023iseedeadpeople} as transfer (gray) and the one ~\cite{zhao2023onevaluating} as transfer+query (gray). To simulate a black-box environment, we adapted these gray-box attacks by employing the CLIP model’s image encoder in lieu of the target model’s encoder, resulting in “transfer (black)” and “transfer+query (black)” variants. As depicted in Table \ref{tab:total}, adversarial samples generated by the original gray-box attacks exhibit a marked increase in textual similarity to the target text when compared to clean samples. Conversely, the black-box adaptations maintain a similarity level akin to that of clean samples, indicating a significant loss of attack capability upon changing the image encoder. This underscores the dependency of gray-box attacks on the target model’s image encoder. Our proposed method \textit{AAA} demonstrates superior attack performance in black-box scenarios compared to the existing methods in their native gray-box settings. This is attributed to the semantic loss inherent in existing gray-box attacks, which constrains their attacking potential. It is noteworthy that our work represents the first black-box attack on image-to-text models. So we can only compare our approach with existing gray-box attacks. We have adapted these gray-box attacks into a black-box version solely to demonstrate their ineffectiveness in a black-box scenario.

% We compare the performance of the more practical and state-of-the-art gray-box attacks ~\cite{lapid2023iseedeadpeople,zhao2023onevaluating} on image-to-text models.  We call the gray-box attack ~\cite{lapid2023iseedeadpeople} as transfer (gray), and the gray-box attack ~\cite{zhao2023onevaluating} as transfer+query (gray). We also modify the existing gray-box attacks into black-box by using the CLIP model’s image encoder instead of the target model’s. We call them transfer (black) and transfer+query (black). As shown in Table \ref{tab:total}, compared to clean samples, the output text of adversarial sample of existing gray-box attacks has a higher similarity with the target text, while the similarity of black-box versions of existing attacks is close to that of clean samples. This means that existing gray-box methods can effectively complete targeted attacks in gray-box scenarios, but completely lose the attack ability when the image encoder changes.  This shows that gray-box attacks heavily rely on the image encoder of the target model. Our method \textit{AAA} has stronger attack performance in black-box scenarios than the existing methods in gray-box scenarios, because the existing gray-box methods have semantic loss that limits their attack ability. It is noteworthy that our work represents the first black-box attack on image-to-text models. Consequently, we can only compare our approach with existing grey-box attacks. We have adapted these grey-box attacks into a black-box version solely to demonstrate their ineffectiveness in a black-box scenario. 

\paragraph{Ablation experiment of our black-box attack.}

We conducted ablation experiments on our \textit{AAA} method. \textit{AAA} (w/o \textit{Attend}) means no attention heatmap to reduce the search space, but the proportional reduction of the search range. \textit{AAA} (w/o \textit{Ask}) means the target text is not from the target semantic dictionary, but random words. Table \ref{tab:total}  shows that losing any module decreases our attack performance. 
In addition, \textit{Ask} performs worse than \textit{AAA} (w/o \textit{Attend}), indicating that finding a target text with lower search difficulty contributes relatively more to the performance of our targeted attack.

\begin{figure*}[t]
\centering
\includegraphics[width=1.0\columnwidth]{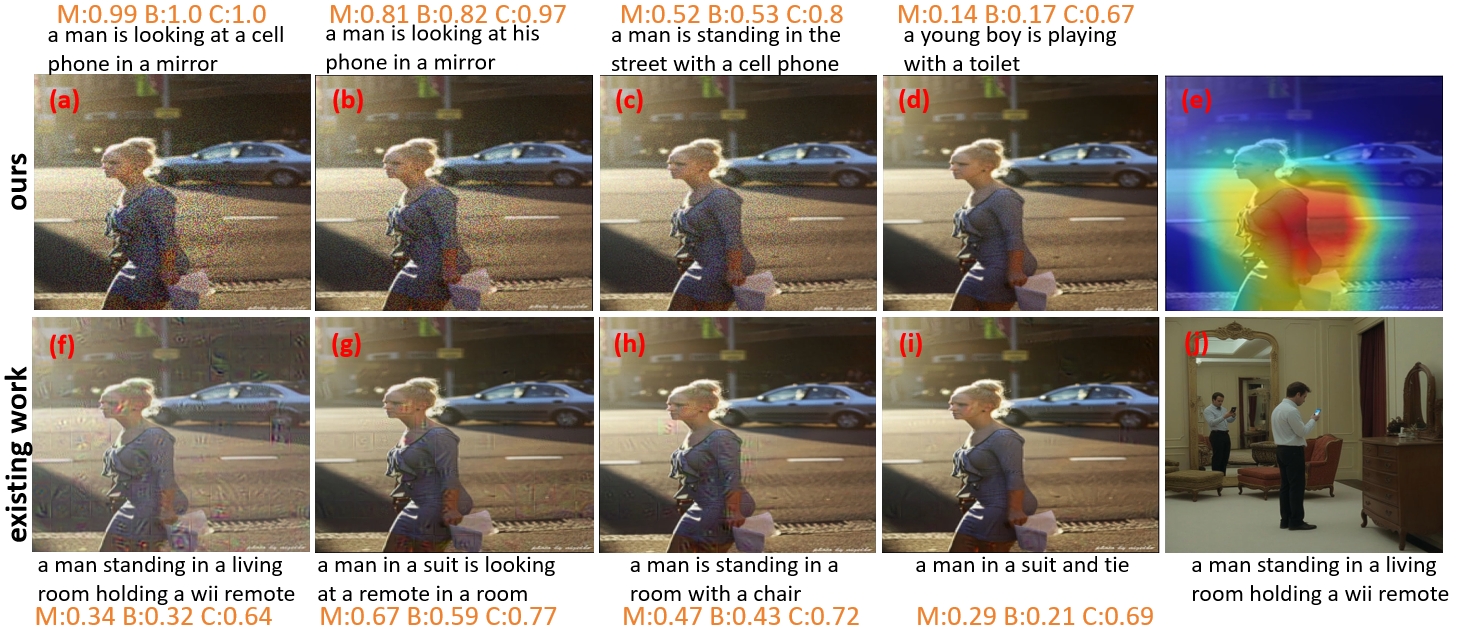}
\caption{Performance of adversarial image attacks varies with perturbation size $\epsilon$. The $\epsilon$ of (a) and (f) is 25, $\epsilon$ of (b) and (g) is 15, $\epsilon$ of (c) and (h) is 10, $\epsilon$ of (d) and (i) is 5. (e) is our attention heatmap of the target text on the image. (j) is the target image generated based on the target text used in existing works. M is METEOR score, B is BLEU score, and C is CLIP score.}
\label{fig:perturbation_size}
\end{figure*}

\paragraph{Qualitative experiment of attention.}
\label{sec: Qualitative experiment of attention}
We presented the optimization curves of \textit{AAA} and \textit{AAA} (w/o \textit{Attend}) in Figure \ref{fig:attention}. Figure \ref{fig:attention} (a) and (b) illustrate the best and average fitness values during \textit{AAA} and \textit{AAA} (w/o \textit{Attend}) optimization of VIT-GPT2 and Show-Attend-Tell. It is evident that the inclusion of \textit{Attend} expedites and enhances the convergence of the population, with an equivalent perturbation size. Consequently, \textit{AAA} exhibits more effective concealment in adversarial perturbations, maintaining the same level of attack efficacy, as depicted in Figures \ref{fig:attention} (d) and (e). Furthermore, we evaluated the impact of selecting different surrogate models during \textit{Attend}. Notably, the sole function of the surrogate model is to compute the attention heatmap. Figure \ref{fig:gradcam_backbone} demonstrates that, despite significant structural variances among several surrogate models, they produce strikingly similar attention heatmaps for the same target text and input images. This similarity arises from mapping the target text to the most pertinent category within the surrogate model’s label space (as Formula \ref{classC}). The position of the same category of objects on the same picture is constant, and the model needs to focus on the object first, no matter what structure it is \cite{wang2021dual}. Performance comparisons, as shown in Figure \ref{fig:gradcam_backbone}, indicate that the similarity in attention heatmaps across different surrogate models leads to similar final attack performances. Therefore, we opted for a stable, well-established, pre-trained model, such as ResNet-50, to serve as our surrogate model.

\paragraph{Qualitative experiment of different perturbation sizes.}

We used the words \textit{mirror}, \textit{cell phone}, \textit{man}, \textit{looking at} from the target semantic dictionary (as shown in Appendix \ref{sec:target semantic dictionary}) to make the target text \textit{a man is looking at a cell phone in a mirror}. We compared output texts of our black-box method \textit{AAA} and the existing gray-box method ~\cite{zhao2023onevaluating} for adversarial samples with different $\epsilon$,  the average pixel perturbation size, in Figure \ref{fig:perturbation_size}. The same conclusion drawn from both methods is that bigger perturbation causes worse concealment and better attack performance; too small perturbation causes attack failure. Moreover, (f) and (j) in Figure \ref{fig:perturbation_size} show that the existing methods have a semantic loss that limits their attack performance. Subjectively, target image (j) accurately draws the semantics of the target text, and the output text of adversarial image (f) perfectly describes the content of the target image (j).  However the adversarial sample (f)'s output text does not have the semantics of the target text. Our method does not have semantic loss, so our black-box method \textit{AAA} does a better targeted attack than the existing gray-box method. More examples of semantic loss are in Appendix \ref{sec:semantic loss example}.

\begin{figure}[t]
\centering
\includegraphics[width=1\columnwidth]{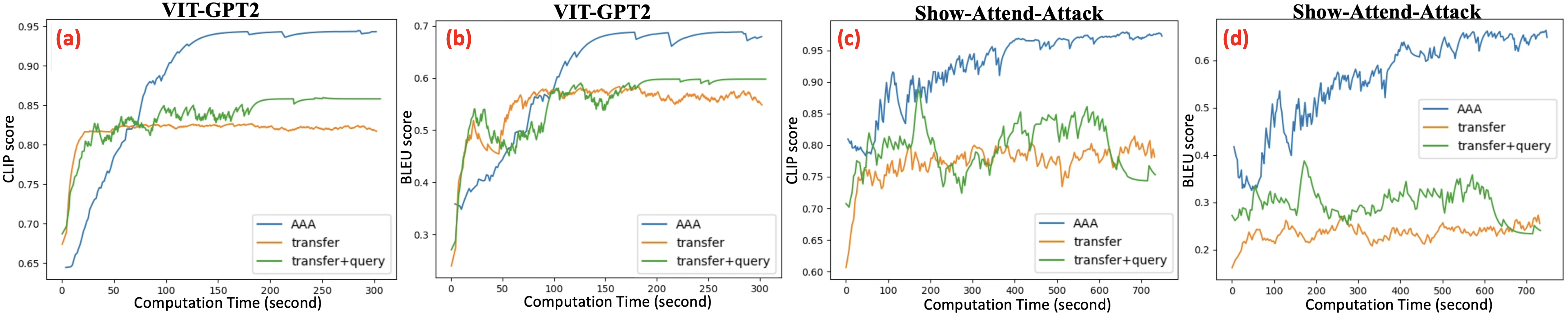}
\caption{Comparison of computation time for generating a single adversarial sample using different adversarial attack methods. The y-axis is a measure of similarity between the generated text and the target text, with higher values indicating better target attack performance. The x-axis represents the computation time, and the shorter the time required to find a stable solution, the better.}
\label{fig:iteration_time}
\end{figure}

\paragraph{Comparison experiment on computation time.}

We evaluated the computational efficiency of various attack methodologies for generating adversarial samples in image-to-text models. As depicted in Figure \ref{fig:iteration_time}, our black-box attack method  \textit{AAA}, demonstrates a longer computation time to reach an optimal solution compared to existing gray-box attacks. For instance, the transfer approach \cite{lapid2023iseedeadpeople} illustrated in Figure \ref{fig:iteration_time} (a) produces an adversarial sample with a CLIP score of 0.82 within a mere 29 seconds, while the transfer+query approach \cite{zhao2023onevaluating} achieves a CLIP score of 0.85 in just 97 seconds. Conversely, our \textit{AAA} method requires 151 seconds to generate an adversarial sample with a superior CLIP score of 0.951. The shorter computation times of the existing gray-box methods are expected due to their ability to access real gradients, which significantly expedites the optimization process. Given that adversarial attacks are not time-sensitive operations and considering that our \textit{AAA} method delivers a more potent attack capability and is applicable in a broader range of realistic black-box scenarios, the trade-off for a higher computational cost is deemed acceptable. Additional experiments on similarity measurements are included in the Appendix ~\ref{sec:meteor_spice}.

% It is worth noting that the reason why we did not choose the number of iterations as the horizontal axis is because the computational cost of one iteration based on evolution and gradient optimization methods is different, so we chose a more intuitive calculation time (in seconds). In addition, more similarity measurement experiments are included in the appendix.

\paragraph{Further analyses.}

\underline{Firstly}, we show the impact of different forms of target semantics $TS$ in \textit{Ask} on the target semantic dictionary, as shown in Appendix \ref{sec:target semantic dictionary}. More ambiguous target semantics can enrich the target semantic dictionary, which also means that the attacker has more choices when designing $y_t$. \underline{Secondly}, we show the effect of different word selection strategies of $y_t$ based on target semantic dictionary on the final attack effect, as shown in Appendix \ref{sec: Word selection strategies}. \underline{Thirdly}, we compare the convergence curves of different population sizes and choose a population size of 40 based on the trade-off of attack performance and convergence efficiency, as shown in Appendix \ref{sec: population size}. \underline{Furthermore}, we compare the effects of different evolutionary algorithms on attack performance and convergence efficiency, as shown in Appendix \ref{sec: optimization algorithms}. \underline{Additionally},  to better observe the attack effect of our framework, we show more examples of attention heatmaps $\textbf{A}$, optimization convergence curves, target text $y_t$, and output text, as shown in Appendix \ref{sec: more example}. \underline{Lastly}, we discuss the limitations of our framework, defense strategies, and future work in Appendix \ref{sec:discussion}.

\section{Conclusion}
In our research, we introduce a novel and practical approach for adversarial attacks on image-to-text models. We propose the \textit{Ask, Attend, Attack} (\textit{AAA}) framework, a decision-based black-box attack method that achieves targeted attacks without semantic loss, even with access limited to the target model's output text. Our framework uses the target semantic directory to guide the creation of target text and attention heatmap to reduce the search space, thereby improving the efficiency of evolutionary algorithms and making our attack harder to detect. Our extensive experiments on the Transformer-based VIT-GPT2 model and the CNN+RNN-based Show-Attend-Tell model demonstrate that our decision-based black-box method outperforms existing gray-box methods in targeted attack performance. These findings highlight the vulnerabilities in current image-to-text models and underscore the need for more robust defense mechanisms, significantly contributing to the field of adversarial machine learning and enhancing the security of vision-language systems.

\bibliographystyle{IEEEtran}
\bibliography{IEEEabrv, reference}

\begin{thebibliography}{10}

\bibitem{li2022blip}
Junnan Li, Dongxu Li, Caiming Xiong, and Steven Hoi.
\newblock Blip: Bootstrapping language-image pre-training for unified vision-language understanding and generation.
\newblock In {\em Proceedings of the International Conference on Machine Learning (ICML)}, pages 12888--12900. PMLR, 2022.

\bibitem{li2023blip2}
Junnan Li, Dongxu Li, Silvio Savarese, and Steven Hoi.
\newblock Blip-2: Bootstrapping language-image pre-training with frozen image encoders and large language models.
\newblock {\em arXiv preprint arXiv:2301.12597}, 2023.

\bibitem{antol2015vqa}
Stanislaw Antol, Aishwarya Agrawal, Jiasen Lu, Margaret Mitchell, Dhruv Batra, C~Lawrence Zitnick, and Devi Parikh.
\newblock Vqa: Visual question answering.
\newblock In {\em Proceedings of the IEEE/CVF International Conference on Computer Vision (ICCV)}, pages 2425--2433. IEEE, 2015.

\bibitem{kim2021vilt}
Wonjae Kim, Bokyung Son, and Ildoo Kim.
\newblock Vilt: Vision-and-language transformer without convolution or region supervision.
\newblock In {\em Proceedings of the International Conference on Machine Learning (ICML)}, pages 5583--5594. PMLR, 2021.

\bibitem{cao2018vggface2}
Qiong Cao, Li~Shen, Weidi Xie, Omkar~M Parkhi, and Andrew Zisserman.
\newblock Vggface2: A dataset for recognising faces across pose and age.
\newblock In {\em Proceedings of the IEEE International Conference on Automatic Face \& Gesture Recognition (FG)}, pages 67--74. IEEE, 2018.

\bibitem{lu2019vilbert}
Jiasen Lu, Dhruv Batra, Devi Parikh, and Stefan Lee.
\newblock Vilbert: Pretraining task-agnostic visiolinguistic representations for vision-and-language tasks.
\newblock {\em Proceedings of the Neural Information Processing Systems (NIPS)}, 32, 2019.

\bibitem{chen2018attacking}
Hongge Chen, Huan Zhang, Pinyu Chen, Jinfeng Yi, and Cho-Jui Hsieh.
\newblock Attacking visual language grounding with adversarial examples: A case study on neural image captioning.
\newblock In {\em Proceedings of the Annual Meeting of the Association for Computational Linguistics}, pages 2587--2597. Association for Computational Linguistics, 2018.

\bibitem{lapid2023iseedeadpeople}
Raz Lapid and Moshe Sipper.
\newblock I see dead people: Gray-box adversarial attack on image-to-text models.
\newblock In {\em Proceedings of the European Conference on Machine Learning and Principles and Practice of Knowledge Discovery in Databases (ECML-PKDD)}, 2023.

\bibitem{zhao2023onevaluating}
Yunqing Zhao, Tianyu Pang, Chao Du, Xiao Yang, Chongxuan Li, Ngai-Man Cheung, and Min Lin.
\newblock On evaluating adversarial robustness of large vision-language models.
\newblock In {\em Proceedings of the Neural Information Processing Systems (NIPS)}, 2023.

\bibitem{kwon2022restricted}
Hyun Kwon and SungHwan Kim.
\newblock Restricted-area adversarial example attack for image captioning model.
\newblock In {\em Wireless Communications and Mobile Computing (WCMC)}. Hindawi, 2022.

\bibitem{bhattad2020unrestricted}
Anand Bhattad, Minjin Chong, Kaizhao Liang, Bo~Li, and D.~A. Forsyth.
\newblock Unrestricted adversarial examples via semantic manipulation.
\newblock In {\em Proceedings of the International Conference on Learning Representations (ICLR)}. ICLR, 2020.

\bibitem{dong2019efficient}
Yinpeng Dong, Hang Su, Baoyuan Wu, Zhifeng Li, Wei Liu, Tong Zhang, and Jun Zhu.
\newblock Efficient decision-based black-box adversarial attacks on face recognition.
\newblock 2019.

\bibitem{shi2022query}
Yucheng Shi, Yahong Han, Qinghua Hu, Yi~Yang, and Qi~Tian.
\newblock Query-efficient black-box adversarial attack with customized iteration and sampling.
\newblock {\em IEEE Transactions on Pattern Analysis and Machine Intelligence}, 45(2), 2022.

\bibitem{jia2021iou}
Shuai Jia, Yibing Song, Chao Ma, and Xiaokang Yang.
\newblock Iou attack: Towards temporally coherent black-box adversarial attack for visual object tracking.
\newblock In {\em Proceedings of the IEEE/CVF Conference on Computer Vision and Pattern Recognition (CVPR)}, 2021.

\bibitem{jiang2023efficient}
Kaixun Jiang, Zhaoyu Chen, Hao Huang, Jiafeng Wang, Dingkang Yang, Bo~Li, Yan Wang, and Wenqiang Zhang.
\newblock Efficient decision-based black-box patch attacks on video recognition.
\newblock In {\em Proceedings of the IEEE/CVF International Conference on Computer Vision (ICCV)}, pages 4379--4389, 2023.

\bibitem{wu2022learning}
Hanjie Wu, Yongtuo Liu, Hongmin Cai, and Shengfeng He.
\newblock Learning transferable perturbations for image captioning.
\newblock {\em ACM Transactions on Multimedia Computing, Communications and Applications}, 18(2), 2022.

\bibitem{omidvar2013cooperative}
Mohammad~Nabi Omidvar, Xiaodong Li, and Yi~Mei.
\newblock Cooperative co-evolution with differential grouping for large scale optimization.
\newblock {\em IEEE Transactions on evolutionary computation}, 18(3):378--393, 2013.

\bibitem{9552479}
Zhenzhong Wang, Haokai Hong, Kai Ye, Guangen Zhang, Min Jiang, and Kay~Chen Tan.
\newblock Manifold interpolation for large-scale multiobjective optimization via generative adversarial networks.
\newblock {\em IEEE Transactions on Neural Networks and Learning Systems}, 34(8):4631--4645, 2023.

\bibitem{hong2023improving}
Haokai Hong, Min Jiang, and Gary~G Yen.
\newblock Improving performance insensitivity of large-scale multiobjective optimization via monte carlo tree search.
\newblock {\em IEEE Transactions on Cybernetics}, 2023.

\bibitem{xu2018fooling}
Xiaojun Xu, Xinyun Chen, Chang Liu, Anna Rohrbach, Trevor Darrell, and Dawn Song.
\newblock Fooling vision and language models despite localization and attention mechanism.
\newblock In {\em Proceedings of the IEEE/CVF Conference on Computer Vision and Pattern Recognition (CVPR)}. IEEE, 2018.

\bibitem{zhang2020fooled}
Shaofeng Zhang, Zheng Wang, Xing Xu, Xiang Guan, and Yang Yang.
\newblock Fooled by imagination: Adversarial attack to image captioning via perturbation in complex domain.
\newblock In {\em Proceedings of the IEEE International Conference on Multimedia and Expo (ICME)}. IEEE, 2020.

\bibitem{ji2020attacking}
Jiayi Ji, Xiaoshuai Sun, Yiyi Zhou, Rongrong Ji, and Fuhai Chen.
\newblock Attacking image captioning towards accuracy-preserving target words removal.
\newblock In {\em Proceedings of the ACM International Conference on Multimedia (ACM MM)}. ACM, 2020.

\bibitem{huang2021igseg}
Qingbao Huang, Chuan Huang, Linzhang Mo, Jielong Wei, Yi~Cai, Ho-fung Leung, and Qing Li.
\newblock Igseg: Image-guided story ending generation.
\newblock In {\em Findings of the ACL: International Journal of Conference on Natural Language Processing}, 2021.

\bibitem{chaturvedi2020mimicandfool}
Akshay Chaturvedi and Utpal Garain.
\newblock Mimic and fool: A task-agnostic adversarial attack.
\newblock {\em IEEE Transactions on Neural Networks and Learning Systems}, 32(4):1801--1808, 2020.

\bibitem{nesterov2017random}
Yurii Nesterov and Vladimir Spokoiny.
\newblock Random gradient-free minimization of convex functions.
\newblock {\em Foundations of Computational Mathematics}, 17:527--566, 2017.

\bibitem{banerjee2005meteor}
Satanjeev Banerjee and Alon Lavie.
\newblock Meteor: An automatic metric for mt evaluation with improved correlation with human judgments.
\newblock In {\em Proceedings of the ACL Workshop on Intrinsic and Extrinsic Evaluation Measures for Machine Translation and/or Summarization (ACL WIEEMMTS)}, pages 65--72, 2005.

\bibitem{selvaraju2017grad}
Ramprasaath~R. Selvaraju, Michael Cogswell, and Abhishek Das.
\newblock Grad-cam: Visual explanations from deep networks via gradient-based localization.
\newblock In {\em Proceedings of the IEEE/CVF International Conference on Computer Vision (ICCV)}, pages 618--626, 2017.

\bibitem{wang2021dual}
Jiakai Wang, Aishan Liu, Zixin Yin, Shunchang Liu, Shiyu Tang, and Xianglong Liu.
\newblock Dual attention suppression attack: Generate adversarial camouflage in physical world.
\newblock In {\em Proceedings of the IEEE/CVF Conference on Computer Vision and Pattern Recognition (CVPR)}, pages 8565--8574, 2021.

\bibitem{zhang2009jade}
Jingqiao Zhang and Arthur~C Sanderson.
\newblock Jade: adaptive differential evolution with optional external archive.
\newblock {\em IEEE Transactions on evolutionary computation}, 13(5):945--958, 2009.

\bibitem{jiang2019semantic}
Jyun-Yu Jiang, Mingyang Zhang, Cheng Li, Michael Bendersky, Nadav Golbandi, and Marc Najork.
\newblock Semantic text matching for long-form documents.
\newblock In {\em Proceedings of the World Wide Web Conference (WWW)}, pages 795--806, 2019.

\bibitem{nlpconnect2022}
NLP Connect.
\newblock vit-gpt2-image-captioning (revision 0e334c7).
\newblock \url{https://huggingface.co/nlpconnect/vit-gpt2-image-captioning}, 2022.

\bibitem{Show2015}
Kelvin Xu, Jimmy Ba, and Kiros Jamie.
\newblock Show, attend and tell: Neural image caption generation with visual attention.
\newblock In {\em Proceedings of the International Conference on Machine Learning (ICML)}, pages 2048--2057. PMLR, 2015.

\bibitem{papineni2002bleu}
Kishore Papineni and Salim Roukos.
\newblock Bleu: a method for automatic evaluation of machine translation.
\newblock In {\em Proceedings of the Annual Meeting of the Association for Computational Linguistics}, pages 311--318, 2002.

\bibitem{radford2021learning}
Alec Radford, Jong~Wook Kim, Chris Hallacy, Aditya Ramesh, Gabriel Goh, Sandhini Agarwal, Girish Sastry, Amanda Askell, Pamela Mishkin, Jack Clark, Gretchen Krueger, and Ilya Sutskever.
\newblock Learning transferable visual models from natural language supervision.
\newblock In {\em Proceedings of the International Conference on Machine Learning (ICML)}, pages 8748--8763. PMLR, 2021.

\bibitem{anderson2016spice}
P.~Anderson, B.~Fernando, M.~Johnson, and S.~Gould.
\newblock Spice: Semantic propositional image caption evaluation.
\newblock In {\em Proceedings of the European Conference on Computer Vision (ECCV)}, pages 382--398. Springer, 2016.

\bibitem{ActionableExplanations}
Zhenzhong Wang, Qingyuan Zeng, Wanyu Lin, Min Jiang, and Kaychen Tan.
\newblock Generating diagnostic and actionable explanations for fair graph neural networks.
\newblock In {\em Proceedings of the Association for the Advancement of Artificial Intelligence (AAAI)}, 2024.

\bibitem{storn1997differential}
Rainer Storn and Kenneth Price.
\newblock Differential evolution–a simple and efficient heuristic for global optimization over continuous spaces.
\newblock {\em Journal of global optimization}, 11:341--359, 1997.

\bibitem{khatib1998stud}
Wael Khatib and Peter~J. Fleming.
\newblock The stud ga: a mini revolution?
\newblock In {\em Proceedings of the International Conference on Parallel Problem Solving from Nature (PPSN)}, pages 683--691. Springer Berlin Heidelberg, 1998.

\end{thebibliography}

\end{document}